# Toyota Smarthome Untrimmed: Real-World Untrimmed Videos for Activity Detection

Rui Dai, Srijan Das, Saurav Sharma, Luca Minciullo, Lorenzo Garattoni, Francois Bremond, Gianpiero Francesca

**Abstract**—Designing activity detection systems that can be successfully deployed in daily-living environments requires datasets that pose the challenges typical of real-world scenarios. In this paper, we introduce a new untrimmed daily-living dataset that features several real-world challenges: Toyota Smarthome Untrimmed (TSU). TSU contains a wide variety of activities performed in a spontaneous manner. The dataset contains dense annotations including elementary, composite activities and activities involving interactions with objects. We provide an analysis of the real-world challenges featured by our dataset, highlighting the open issues for detection algorithms. We show that current state-of-the-art methods fail to achieve satisfactory performance on the TSU dataset. Therefore, we propose a new baseline method for activity detection to tackle the novel challenges provided by our dataset. This method leverages one modality (i.e. optic flow) to generate the attention weights to guide another modality (i.e RGB) to better detect the activity boundaries. This is particularly beneficial to detect activities characterized by high temporal variance. We show that the method we propose outperforms state-of-the-art methods on TSU and on another popular challenging dataset, Charades.

✦

## 1 INTRODUCTION

ACCORDING to a recent report of the United Nations [1], the global population aged 60+ is projected to grow from 0.9 billion in 2015 to 1.4 billion in 2030. This demographic trend translates to the dramatic need for an increase of the workforce in healthcare. A great support to the healthcare workforce could come from activity detection systems, which help monitor the health state of older patients and support the early detection of potential physical or mental disorders. For instance, monitoring patient eating habits allows doctors to track the state of a patient and react before serious health conditions arise. Thanks to such systems, seniors could stay longer at home without the need of being hospitalized, which would greatly improve their comfort and quality of life. Building such monitoring systems requires fine-grained understanding of long untrimmed videos.

In recent years, numerous datasets for activity classification in trimmed videos have been proposed [2], [3], [4], whereas very little has been done for activity detection in untrimmed videos. By activity detection, we mean predicting the activity label as well as the temporal boundaries within an input video. This detection task has to cope with important open challenges: i) handling the combinatorial explosion of activity proposals while detecting accurate temporal boundaries in long video sequences, ii) managing concurrent activities, and iii) distinguishing between background and foreground activities (e.g. *standing still/using telephone*). In this work, we focus on untrimmed videos of Activities of Daily Living (ADLs). These videos contain

- R. Dai, S. Das, S. Sharma and F. Bremond are with the Inria and Université Côte d'Azur, 2004 Route des Lucioles, 06902 Valbonne, France. E-mail: {rui.dai, srijan.das, saurav.sharma, francois.bremond}@inria.fr
- L. Minciullo, L. Garattoni . and G. Francesca are with Toyota Motor Europe, Hoge Wei 33, B - 1930 Zaventem, Belgium.

activities that usually occur in the daily lives of older people. Typically ADLs feature activities with similar motion (e.g. *eating/drinking*), activities with high temporal variance (e.g. *putting on glasses* in 5 sec./ *reading* for 10 min.), or subtle motions (e.g. *stirring the coffee*).

Most of the untrimmed video datasets that are widely adopted in the literature do not focus on ADL. These datasets are often collections of videos from the web [5], [6], [7], [8], [9], [10]. For instance, ActivityNet [5] and Multi-Thumos [9] are collections of a large number of videos encompassing sports and outdoor activities. These activities are often characterized by high inter-class variation due to large and distinctive motions. Other datasets contain movie excerpts or instructional videos [11], [12]. The videos in these datasets retain only the key part of the activity and are mostly recorded by a cameraman from a frontal viewpoint, with nearly no occlusions.

Some ADL datasets have been proposed in the past few years [13], [14], [15]. These datasets share common characteristics: i) Subjects usually follow a rigid script, which results into unnatural movements; ii) Videos and thus activities are usually short; iii) Subjects are usually centered in the middle of the frame and perform activities facing the camera (i.e. high camera framing). These characteristics do not reflect the spontaneity of human activities in real-world scenarios.

Motivated by the shortcomings of current datasets, we introduce Toyota Smarthome Untrimmed (TSU). TSU provides realistic untrimmed videos with diverse spontaneous human activities and real-world settings. We invited 18 volunteers to the recording session in a smarthome. The volunteers are senior people in the age range of 60 to 80 years. Each volunteer was recorded for 8 hours in one day. The resulting data consists of 536 long RGB+D videos with 51 annotated activity classes. This dataset is an extension of the previously published Toyota Smarthome dataset [16], which is designed for the classification task of clipped videos. Unlike previous datasets, these videos are unscripted. Activities are annotated with both coarse and fine-grained labels. The dataset poses several challenges: high intra-class temporal variance, high class imbalance, composite and elementary activities, and activities with similar motion. In our data acquisition process, each participant was recorded continuously for 8 hours. We believe that this setup reduced camera awareness in the participants, leading to increased spontaneity. Consequently, in TSU, the participants may commit errors, search for items and repeat several times the same activity before succeeding. The fact that activities are performed in a spontaneous manner also amplifies other challenges such as low camera framing and high temporal variance. Some of the challenges in TSU dataset are illustrated in Fig. 1. In section 3, we analyse in detail the characteristics and novelty of the proposed dataset.

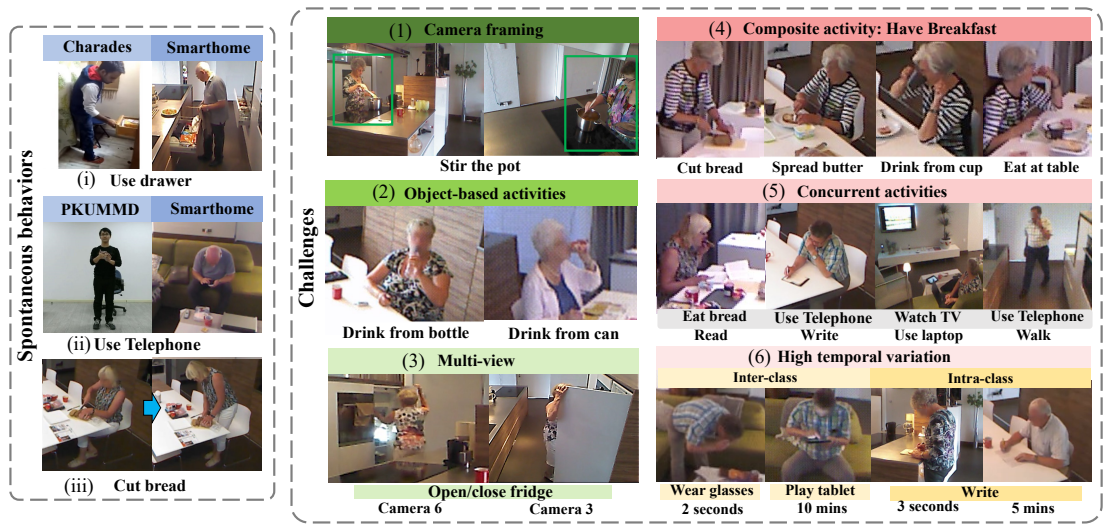

Fig. 1: Overview of the challenges in TSU. On the left part, we present challenges related to **spontaneous behaviours**: For the first two examples, we present the activity following a strict script on the left, and the same activity performed spontaneously in TSU on the right: i) Different from *using drawer* performed quickly, once per video [13], in TSU, *using drawer* may be repeated several times in a video, and the subject may keep several drawers open at the same time to facilitate finding things. ii) In [14], subject uses shortly the *telephone* while looking at the camera. In contrast in TSU, the subject is deeply involved with his *telephone* and the activity may last several minutes instead of few seconds. iii) In TSU, subject may stayed seated or *stand up* to *cut the bread* in an easier manner. Besides the spontaneous behaviours, we also illustrate on the right part the following real-world challenges: (1) Camera framing: subject is not in the middle of the image and can be even outside the field of view. (2) Object-based activities: similar activities can be performed while interacting with different objects. (3) Multi-views: activities look different in appearance from different view points. (4) Composite activity: composite activities can be split into several elementary activities (e.g. While *having breakfast*, we may *cut bread*, *spread butter* and *eat at the table*). Moreover, these complex composite activities can last a long period of time. Large variations of appearance make the recognition challenging, requiring to understand the composition of elementary activities to better recognize the composite activities. (5) Concurrent activities: activities can be performed concurrently (e.g. *take note* while *having a phone call*). (6) High temporal variation: in the same untrimmed video, we may have related short activities (e.g. *taking on glasses*) and long activities (e.g. *playing tablet*). Different instances of the same activity class can also be short or long (e.g. *writing*) corresponding to high intra-class temporal variance.

Experimentally, we find out that state-of-the-art activity detection methods fail to address the aforementioned real-world challenges offered by TSU. We also find that the model trained on the untrimmed TSU outperforms the same model trained on the trimmed version [16], reflecting the difficulty of handling background actions. So, the question remains, how to address these real-world challenges for the task of activity detection? To this end, we design a novel activity detection baseline method, Attention Guided Net (AGNet), which builds upon existing temporal convolutional networks. This method uses two input modality streams (e.g. RGB and 3D Poses). The attention module generates the attention map from one stream to guide the other stream to predict more precise activity boundaries.

In general, the low performance achieved by activity detection methods on TSU highlights the many challenges that are yet to be addressed. To promote the development of novel activity-detection methods that can better address such challenges, we are releasing TSU to the research community.

## 2 RELATED WORK

In this section, we give an overview of publicly available untrimmed activity detection datasets and related state-of-the-art algorithms.

TABLE 1: Untrimmed dataset comparison along the seven real-world challenges. *Indicates that the activity labels are provided in terms of caption. With 'woT' we indicate that the composite labels are provided without the corresponding temporal boundaries.

| Dataset | Spontaneous behaviour | Camera framing | Object-based activities | Multi-view | Composite activities | Concurrent activities | Var. activity duration | Temporal annotation | View type | Video type |
|---|---|---|---|---|---|---|---|---|---|---|
| MEVA [17] | High | Low | Yes | No | No | No | Low | Precise | Monitoring | Surveillance |
| ACTEV/VIRAT [18] | High | Low | Yes | No | No | No | Low | Precise | Monitoring | Surveillance |
| DALY [8] | Medium | High | No | No | No | Yes | Low | Precise | Shooting | Web |
| HACS [10] | Medium | High | Yes | No | No | No | Medium | Precise | Shooting | Web |
| YouTube'8M-Segments [7] | Medium | High | No | No | No | No | Low | Noisy | Shooting | Web |
| ActivityNet-200 [5] | Medium | High | Yes | No | No | Few | Medium | Precise | Shooting | Web |
| Thumos14 [6] | Medium | High | No | No | No | No | Low | Precise | Shooting | Web |
| Multi-Thumos [9] | Medium | High | No | No | No | Yes | Medium | Precise | Shooting | Web |
| AVA [11] | Medium | High | No | No | No | Yes | Low | Precise | Shooting | Movie |
| How2 [19] | Low | High | Yes* | No | - | - | - | Noisy | Shooting | Instructional |
| HowTo100M [12] | Low | High | Yes* | No | - | - | - | Noisy | Shooting | Instructional |
| Coin [20] | Low | High | Yes | No | No | No | Medium | Noisy | Shooting | Instructional |
| ADL [21] | Medium | High | Yes | No | No | No | Low | Precise | Egocentric | ADL |
| Charades-ego [22] | Medium | High | Yes | No | No | Yes | Low | Precise | Egocentric | ADL |
| 50 Salades [23] | Medium | High | Yes | No | No | No | Low | Precise | Top-view | Cooking |
| EGTEA Gaze+ [24] | Medium | High | Yes | No | No | No | Low | Precise | Egocentric | Cooking |
| EPIC-KITCHENS [25], [26] | High | High | Yes | No | Few | Few | High | Noisy | Egocentric | Cooking |
| MPII Cooking 2 [27] | Low | High | Yes | No | woT | No | Medium | Precise | Shooting | Cooking |
| Breakfast [28] | Medium | Medium | Yes | Yes | woT | No | Medium | Precise | Shooting | Cooking |
| CAD-120 [29] | Low | High | Yes | No | Yes | No | Low | Precise | Shooting | ADL |
| DAHLIA [15] | High | Low | No | Yes | No | No | High | Precise | Monitoring | ADL |
| PKU-MMD [14] | Low | High | No | Yes | No | No | Low | Precise | Shooting | ADL |
| Charades [13] | Low | High | Yes | No | No | Yes | Low | Precise | Shooting | ADL |
| Toyota Smarthome Untrimmed | High | Low | Yes | Yes | Yes | Yes | High | Precise | Monitoring | ADL |

## 2.1 Activity detection datasets

The availability of videos replicating real-world challenges is crucial to design robust activity detection algorithms. Among existing datasets, only few of these challenges are properly addressed. To understand the limitations of currently available datasets, we introduce the following 7 real-world challenges.

**Spontaneous behaviour**: activities in the real-world are performed naturally. However, most existing datasets are acquired by providing the subjects with a strict script. Besides, as the subjects are aware that their activities are being recorded, they often overact. To quantify spontaneous behaviour, we define a heuristic that considers three aspects: (i) Scripted or unscripted: The datasets following a strict script always have lower spontaneity. We assign the datasets that follow strict script 1 point; the datasets following a coarse script (e.g. cooking a specific meal in a video) 0.5 point; unscripted 0 point. (ii) Camera Awareness: Camera awareness also affects spontaneity. We assign 1 point to the datasets recorded by the cameraman/self-recorded/wearable sensor. We assign 0.5 point to datasets with continuous videos that were recorded for a long duration (at least 30 minutes). For monitoring datasets recorded for a long duration, we assign 0 point. (iii) Environment: it is also an important factor for spontaneity. Activities are often more spontaneous when performed in a familiar environment. Here, we assign a dataset that is recorded in an unfamiliar location 1 point, in a familiar location (e.g. home) 0 point. Datasets with continuous videos that were recorded for a long duration in the same environment are given 0.5 point, as people get accustomed to the location. Following these criteria, we re-evaluate all datasets. The datasets with less than 1 point are considered as featuring high spontaneity, more than 2 points obtained low spontaneity, the others are rated with medium spontaneity. **Camera framing**: when videos are recorded by a cameraman, subjects mostly appear in the middle of the image and facing the camera (high camera framing). On the other hand, when videos are recorded automatically by a monitoring system using fixed cameras, subjects can often be offset from the center, occluded or partially outside the field of view (low camera framing). **Object-based activities**: similar activities that can be performed while interacting with different objects (e.g. *drinking from cup* or *from bottle*) are more challenging to classify. In Table 1, *object-based activities* indicates the availability of object level fine-grained annotation for these activities. **Multi-views**: activity detection methods need to be robust against view-point variations. Therefore, benchmark datasets should provide samples of the same activities recorded from different views. **Composite activities**: Some complex ADLs can be decomposed into several elementary activities. For example, *having breakfast* may contain elementary activities like *cutting bread*, *spreading butter* and *eating at table*. In Table 1, the *composite activities* column indicates whether the dataset provides annotation for both composite activities and their respective elementary activities. **Concurrent activities**: activities, such as *making a phone call* and *taking notes* may be performed simultaneously. The appearance of activities can drastically change when multiple activities are performed at the same time. In Table 1, *concurrent activities* indicates whether the dataset provides samples and annotations in which activities are performed simultaneously. **Variation of activity duration**: this property indicates the level of variation in the length of activities in the dataset. In this table, the high variation indicates that the average duration of an activity class is more than 80 times larger than the one of the lowest activity class. The low variation indicates that the highest average duration of an activity class is less than 30 times than the one of the lowest activity class.

To be noted that, activity detection methods need precise temporal annotation (i.e. start time and end time) for each activity. We consider that a dataset features Noisy annotation when: (i) the dataset provides temporal annotation only for part of the activities in the video [7], or (ii) the dataset is coarsely annotated by the audio [25], or iii) the dataset only provides caption of the video [12], [19].

Table 1 summarizes the comparison of most used public untrimmed video datasets based on the above challenges.

Below, we detail how these untrimmed datasets differ from our proposed TSU.

### 2.1.1 Surveillance datasets

Surveillance datasets, such as VIRAT and MEVA [17], [18], have fixed camera views and are designed to monitor human activities in the wild. These datasets are collected in natural scenes showing people performing normal activities in standard contexts, most of the time outdoors. Besides, activities look natural as they are performed by actors following a light script. For these datasets, only few simple human activities are annotated (i.e. crouching, standing...) along with the object information (i.e. carrying a box). These datasets thus differ from TSU as the complexity of surveillance activities is significantly lower than the one from daily-living activities, for example, no concurrent & composite activities.

### 2.1.2 Web & Youtube & Movie datasets

A large number of datasets are collected from YouTube or movies [5], [6], [7], [8], [9], [10], [11]. Most of these videos are self-recorded or recorded by a cameraman from a single view, which causes the subject to be centered within the image frame (i.e. high camera framing), facing the camera and with limited occlusions. These videos are carefully selected and only the key parts of the activities are retained, in which the subjects always perform the activities smoothly without hesitation in front of the camera (i.e. reduced spontaneity). Thus, these videos are less representative of real-world scenarios compared to TSU videos.

### 2.1.3 Instructional videos

Similar to the above category of datasets, instructional videos [12], [19], [20] are collected from internet sources. These videos provide intuitive visual examples for learners to acquire knowledge to accomplish different tasks. In contrast to TSU, these instructional videos have noisy annotations which are often text descriptions [12], [19] and follow strict temporal ordering of the activities [20]. Similar to web videos, the subjects always perform the activities smoothly without hesitation in front of the camera [12], [20]. These characterizations of the instructional videos are not adequate for real-world activity detection task.

### 2.1.4 Activities of daily living (ADL)

Activities of daily living are performed in indoor environments such as homes or labs. These activities are usually characterized by low inter-class variation and subtle motion. Below, we discuss the ADL datasets categorized by their camera viewing angle.

**Egocentric view datasets:** In Egocentric datasets [21], [22], [24], [25], [26], the videos are recorded with a wearable camera (i.e. reduced spontaneity) or from a top view [23] that captures the scene directly in front of the user at all times, in which only hands are visible in the center of the camera view (i.e. high camera framing). Egocentric videos are designed to study the activities, where the user's hands are manipulating various objects. However, the egocentric paradigm can only collect the activity information from a very restricted viewpoint. This restricted viewpoint makes the appearance of egocentric activities very different from third person view datasets like TSU (e.g. poses are unavailable) and prevent the recording of those activities that cannot be observed from this viewpoint (e.g. *making a phone call*). Due to these characteristics, our comparison mainly focuses on third-view datasets.

**Third person view datasets:** Many of the ADL datasets [27], [28], [30] are limited to kitchen activities. **MPII Cooking 2** [27] and **Breakfast** [28] contain only cooking activities (like *preparing recipes or making breakfast*). The subject is asked to cook a single dish (i.e. composite activities) in each video in these datasets. However, there are no temporal boundaries (i.e. no ground truth timestamps) for the composite activities as they correspond to a whole video. Besides, some of the composite activities occur only once in the dataset. In these datasets, subjects are asked to prepare a specific recipe in a video clip, therefore the activities are performed in rapid succession without hesitation or mistake (reduced spontaneity). Moreover, the dataset lacks the presence of secondary activities irrelevant to *cooking* (e.g. *drinking water*) but often occurs in real-life. In **MPII Cooking 2**, the subjects follow strict scripts (i.e. low spontaneity) and are always in the center of the frame (i.e. high camera framing). Although the videos are recorded from 8 camera views, only a single view is released for this dataset. In **Breakfast**, the hands of the subjects and the objects used are always at the center of the frame without much occlusion (i.e. medium camera framing). Moreover, the number of views are not fixed, even in the same kitchen (from 2 to 5). As mentioned, the subjects in these two datasets perform the activities quickly without much hesitation, which means the datasets are characterized by medium temporal variation and no concurrent activities. So, in the following, we present the datasets that encompass a larger variety of ADLs which are not only restricted to kitchen activities and where not only the top body part can be observed.

**CAD-120** [29] is a small dataset (about 60 K frames in total). This dataset comprises of 20 different activities (including composite and object-based activities) performed by four people in different rooms. The subjects are always in the center of the scene performing short sub-activities following a script (i.e. high camera framing & no spontaneity). Because of the simplicity of activities, current state-of-the-art methods [31], [32] can already achieve excellent results on this dataset. **DAHLIA** [15] is recorded in a single room in a lab with 44 subjects. Each subject has about 40 min recording from 3 fixed camera views (i.e. high spontaneity, low camera framing). The dataset contains only 8 coarse activity classes, thus it does not have the challenges of concurrent, composite and object-based activities. In **PKU-MMD** [14], the videos are recorded from 3 camera views. The activities are performed in the center of the scene by the subjects following a strict script. Besides, there are pauses in between the activities which makes the problem of distinguishing between an activity and background easier compared to real-world scenarios. Thus, this dataset lacks spontaneity & concurrent activities in addition to high camera framing. **Charades** [13] explores object-based activities and concurrent activities. The videos are recorded by hundreds of people in their private homes following strict scripts. Although Charades depicts large numbers of environment diversity, these self-recorded activities are very short (30 sec./video, 10 sec./activity) with low variation of activity duration and in general performed in unnatural manner (overacted), in the center of the camera view (high camera framing). All in all, current ADL datasets address only partially the 7 aforementioned challenges of real-world scenarios. This motivates us to propose TSU.

## 2.2 Activity detection methods

In this section, we review how previous methods address real-world challenges for activity detection. The task of activity detection involves two steps: (1) Extracting frame or segment level features using a model trained for activity classification, we call this step video encoding; (2) Modeling temporal relations among the previously extracted features for the detection task. Below we present the relevant methods for each of these two steps.

### 2.2.1 Video encoding

Learning video encoding is an important factor for video understanding problems [33], [34]. In this step, frame or segment level features are extracted using a model which is trained on activity clips. These features are further input to a model which is trained for the task of activity detection. Thus, the efficacy of the detection task highly relies on the quality of the extracted features or, in other words, the learned representations of the activity classification models. These classification models vary based on the input data modality. For instance, 3D human poses are generally processed by sequential networks whereas RGB images and optical flow images are generally treated by 3D convolutional networks.

**3D human pose** is a popular modality which provides the location of the key joints of the subject for every frame. Skeleton attracted considerable attention due to their strong adaptability to dynamic motion and complicated background [2], [4]. Conventional deep learning based methods manually structure the skeletons as a sequence of joint-coordinate vectors [2], [35]. However, representing the skeleton data as a vector sequence can not fully express the dependency between correlated joints. Recently, graph convolutional networks (GCNs) have been applied to model the skeleton data. Yan et al. [36] have constructed a spatial graph based on the natural connections of joints in the human body. Inspired by [36], Shi et al. [37] have proposed a two-stream GCN to better model the spatial information within a short period of time. However, skeletons can only represent the pose of the person performing an activity. But what about contextual information like environmental details (e.g. sink for *clean dishes with water*), encoding object information (e.g. glasses)? For that, we need the RGB frames.

**RGB images** are utilized by many effective methods in order to model the appearance information. Few works learn appearance features from frame-level classification of activities, using 2D CNNs [38]. 3D CNNs are the natural evolution of their 2D counterparts. Du et al. [34] have proposed 3D CNNs (C3D) to capture spatio-temporal patterns from a sequence of 8 RGB frames. In the same vein, I3D [33] inflates the kernels of ImageNet pre-trained 2D CNN to jump-start the training of 3D CNNs. While these methods are effective for the recognition of fine-grained and object-based activities with a short temporal extent [39], [40], they are too rigid and computationally expensive to handle minute-long videos. In order to effectively learn temporal localization of activities in long videos, the existing detection methods [41], [42], [43] process the videos on top of the aforementioned 2D or 3D CNNs [33], [34], [38], [44].

### 2.2.2 Temporal modeling

The difference between activity classification and detection is the way long-term and complex temporal relations are processed. After the step of encoding videos, activity detection can be seen as a sequence-to-sequence problem. Recurrent Neural Networks (RNNs) [9], [45], [46] have been popularly used to model the temporal relations between frames. However, they only implicitly capture relationships between certain activities with high motion. Furthermore, due to the vanishing gradient problem, RNN-based models can only capture a limited amount of temporal information and short-term dependencies.

Temporal Convolutional Networks (TCNs) are another group of temporal processing methods. In contrast to RNN-based methods, TCNs can process long videos thanks to the fact that kernels share weights for all the time steps. The result is a feature vector preserving the spatial information, along with contextual information from the neighboring frames. Some recent variants of TCNs for activity detection include Dilated-TCN [41] and MS-TCN [43]. Dilated-TCN [41] increases the temporal reception field by using dilated convolutions to model longer temporal patterns. This is extended by MS-TCN [43], which stacks multiple Dilated-TCNs to construct a multi-stage structure, where each stage refines the prediction of the previous one. However, the dilated-based TCNs [43] have dozens of dilated-convolution layers (e.g. 5×10 layers in MS-TCN) and there is no mechanism to combine the local and global features. Hence, the top layer contains only the information of the high-level reception field, but lacks the local features. Besides, the number of filters for each layer is small and can only process datasets with videos characterized by simple temporal relations [23], [28]. Moreover, the number of filters can not be increased in this deep structure due to the computational cost.

Further, the introduction of datasets like MultiTHU-MOS [9] and Charades [13] with dense labelling and concurrent activities (i.e. multi-labels), pushed more and more methods to attempt modeling complex temporal relations between activity instances. Piergiovanni et al. have proposed a global representation, namely super-event [42]. In this model, Cauchy distribution based filters process the video across time to learn a latent contextual representation of the activities on particular sub-intervals of the video. The set of filters are summed by a soft attention mechanism to form the global super-event features. During prediction, the local I3D features are used along with the super-event features to better handle the composed activities. Similarly, Piergiovanni et al. [47] have introduced Temporal Gaussian Mixture (TGM) layers. In contrast to standard convolution layer, TGM computes the filter weights based on Gaussian distributions, which enables TGM to learn longer temporal structures with a limited number of parameters. Although the above methods achieve state-of-the-art in modeling complex temporal relations, the non-adaptive receptive field limits the ability of the models to capture the dynamics for both short and long patterns.

To summarise, state-of-the-art methods have high performance on ADL datasets containing simple activities such as PKU-MMD and DAHLIA. However, these methods still struggle to get reasonable performance on more complex datasets such as Multi-Thumos and Charades. Thus, we propose a new ADL dataset TSU to explore how these state-of-the-art methods perform in real-world conditions.

## 3 TOYOTA SMARTHOME DATASET

In this section we describe the main features of Toyota Smarthome Untrimmed dataset. Our goal is to create a large scale dataset with daily-living activities performed in spontaneous manner.

### 3.1 Data collection

#### 3.1.1 Collection Setup

We use 7 Microsoft Kinect sensors in the recording phase. The apartment plan and camera locations are shown in Fig. 5. Cameras 1 and 2 cover the dinning room area, 4 and 5 the living room, 3, 6 and 7 the kitchen. Thus, we have a coverage over the entire apartment from at least 2 distinct viewing angles. The videos are recorded at 20 frames per second, the size of RGB is VGA (640×480), the standard resolution in most real-world scenarios. The dataset offers 3 modalities: RGB, depth and 3D skeleton (i.e. pose) (see fig. 2).

For the 3D skeletons, we fine-tune LCR-Net++ [48] on this dataset and then extract the 2D skeletons. Finally these 2D skeletons are processed through VideoPose3D [49] and refined by SSTA-PRS [50] to extract the 3D skeletons. We observe that this mechanism extracts 3D poses of better quality compared to those obtained using depth or LCRNet++.

#### 3.1.2 Data collection protocol

One of the key applications of daily-living activity detection is older patient monitoring. Thus, in our dataset, we invited 18 volunteers to our dataset recording sessions. The age of the volunteers ranges between 60 and 80 years old. Each volunteer was recorded for 8 hours in one day starting from morning at 9 a.m. until afternoon at 5 p.m.. On the day of recording, the volunteer arrived in the apartment at 8 a.m. and had a visit to get acquainted with the place and to learn how to use the household equipment such as coffee machine, television, remote control, etc.. The volunteers also

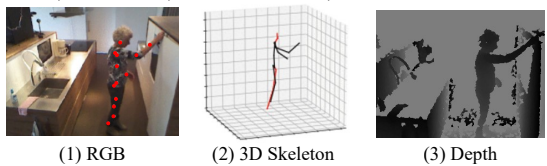

(1) RGB     (2) 3D Skeleton     (3) Depth

Fig. 2: Available modalities in Toyota Smarthome Untrimmed. Note: in the sub-figure of RGB modality, we also mark the 2D skeleton joints.

received an informal description of what it was expected with reference to having meals and interacting with anything in the apartment as it was a normal day at home. No further guidance was provided about how the activities should be performed.

In total, we recorded more than 1000 hours of video data. Based on these data we prepared two datasets: Toyota Smarthome dataset [16], previously published, and Toyota Smarthome Untrimmed dataset that is introduced in this paper.

### 3.2 Toyota Smarthome Trimmed dataset

Toyota Smarthome Trimmed [16] has been designed for the activity classification task. It consists of 16K short RGB+D clips of 31 activity classes. Each clip is about 12.5 sec. long and contains only one activity. Unlike previous datasets [2], [3], activities were performed in a natural manner. As a result, the dataset poses a unique combination of challenges: high intra-class variation, high class imbalance, and activities with similar motion and high duration variance. Activities were annotated with both coarse and fine-grained labels. These characteristics differentiate Toyota Smarthome Trimmed from other datasets for activity classification.

### 3.3 Toyota Smarthome Untrimmed dataset

Toyota Smarthome Untrimmed and Toyota Smarthome Trimmed [16] are obtained from the same recording footage. Different from the Toyota Smarthome Trimmed, TSU is targeting the activity detection task in long untrimmed videos. Therefore, in TSU, we kept the entire recording when the person is visible. The dataset contains 536 videos with an average duration of 21 mins. Since this dataset is based on the same recording as Toyota Smarthome Trimmed version, it features the same challenges and introduces additional ones. In section 3.3.1, we describe the annotation protocol. Then, we present the properties of the TSU dataset in section 3.3.2, we present its challenges in section 3.3.3, and finally we compare this untrimmed version of the dataset (i.e. TSU) with its trimmed version in section 3.3.5.

#### 3.3.1 Annotation protocol

TSU is designed particularly for the activity detection task. With the support of a medical staff, we have identified 51 activities of interest to annotate. A team of annotators manually annotated the videos using the open-source toolkit ELAN [51]. The videos were annotated individually without relying on the fact that some camera views overlap. The annotation process took more than 6 months, including verification and quality checks. We performed the quality check with the help of 5 annotators. We estimated the precision of the annotation by considering the same 50 long videos annotated by different annotators. These 50 videos are randomly chosen and cover all the subjects and camera views. The precision of annotation of those 50 videos is 96.8%. Additionally, we reviewed, normalized and corrected the 25 hours of annotation by checking again the videos where the methods were achieving low activity detection performance. Fig. 3 shows an example of the annotation. This example corresponds to composite activity *cooking*. While *cooking*, the subject abruptly stops cutting vegetables and starts heating water in a pot so that she can have boiled water after cutting the vegetables. After setting up the stove, she resumes cutting the vegetables. This process does not follow a strict temporal order and reflects the spontaneous behaviour of the participant.

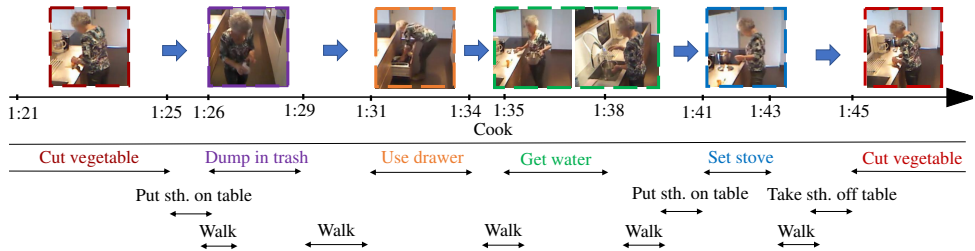

Fig. 3: An example of annotation on TSU dataset. '←' and '→' indicate respectively the start and end of an activity.

### 3.3.2 Dataset Properties

The result of the extensive annotation process is a rich corpus of activities. Fig. 4 presents the diversity of activities in this dataset. The activities are categorized into composite and elementary activities. **Composite activities** are the complex activities that are composed of several **elementary activities** that may or may not follow a temporal ordering. TSU contains 5 composite activities which are relatively long. Elementary activities are atomic activities which may be performed concurrently in time. These activities may or may not be part of a composite activity. TSU contains 46 elementary activities and these activities may be long or short. In Fig. 4 (c), we illustrate the composite activity *cooking*, with its elementary activities. In Fig. 4 (a) and (b), the composite and its corresponding elementary activities are marked with the same color.

TSU contains a rich diversity of elementary activities. We present three challenging scenarios that might occur while attempting to recognize these activities. Firstly, the dataset contains pose-based activities for which poses could be sufficient for classification. In contrast, the appearance information may not improve the recognition of these activities. In Fig. 4 (d), we provide 8 such pose-based activities. For example, *sit down* only needs the 3D poses to be distinguished, whereas the books and laptop around the subject may mislead an appearance-based classifier to recognize an activity related to those objects, such as *reading*. Secondly, TSU contains many elementary activities characterized by similar motions and interactions with objects. These objects provide strong clues to distinguish an activity. However, a reliable detection of the object while processing the whole video is a challenge. Sometimes, the objects are occluded within the hands of the subject, like in the case of *grasping a cup while drinking*. As a result, these activities with similar motion are often miss-classified amongst each other. In Fig. 4 (e), we provide 22 such activities. For example, the subjects performing *use fridge* and *use cupboard* have very similar poses. A fine understanding of the object information (e.g. fridge and cupboard) may facilitate the recognition of these activities. Finally, the dataset contains fine-grained activities characterized by subtle motions, which presents additional challenges for the recognition task. In Fig. 4 (f), we describe 7 such activities. For example, subjects who perform the activity *Stir coffee/tea* move only slightly their wrist and forearm. Compared to activities with pronounced motions, such as *sitting down*, learning discriminative representations for these activities with subtle motions is very challenging.

We further analyze the distribution of the activities in TSU in Fig. 5. We first provide a pictorial representation of the apartment along with the camera placements. TSU features multi-view settings, as all the activities are captured by more than one camera. Then, we provide 6 statistics pertaining to the activity distribution in the dataset. Fig. (a) depicts a distribution of activity instances across the different rooms. Most activities occur in the living room, then kitchen and dinning room. This is similar to real life distribution as we spend most of our time in the living room. Correspondingly, Fig. (f) presents the distribution of environment for each activity. We find that 51% of the activities are environment independent. For instance, we can *eat snack* or *work with laptop* in all these three environments. However, activities that rely on specific equipment occur in the same environment, such as *using oven* in the kitchen. Fig. (b) shows the activity distribution across the activity duration. We find that in TSU, most activities are short activities, followed by medium and long activities. This is because long activities have few occurrences but longer duration. Interestingly, short activities are often more challenging to detect compared to the longer ones [52]. Fig. (c) shows the distribution of activities based on their intra-class temporal variance. We notice that 22% of the activities have high temporal variance (i.e. vary more than 500 sec.). Correspondingly, Fig. (e) provides the heat map of the temporal variance of these activities. The lighter grey means that the temporal variance is higher. Such intra-class variance within the same activity class further complicates the task of detection. Finally, Fig. (d) provides the occurring frequency for every activity in the dataset. We have a non-uniform distribution of activities following the Zipf's law [53]. This long-tail distribution characterizes the real-world scenarios [13], [54].

In addition, we leverage the spatial distribution of the person location to illustrate the camera framing property. We use the key-joint locations of Poses to compute the coordinates of the human position. Fig. 6 shows the spatial distribution of the person center location in different views. Compared to other similar datasets, TSU exhibits a significantly larger spatial scatter for all camera views. In most cases, the subjects move along the edge of the camera coverage area. Consequently, we consider TSU to have relatively low camera framing.

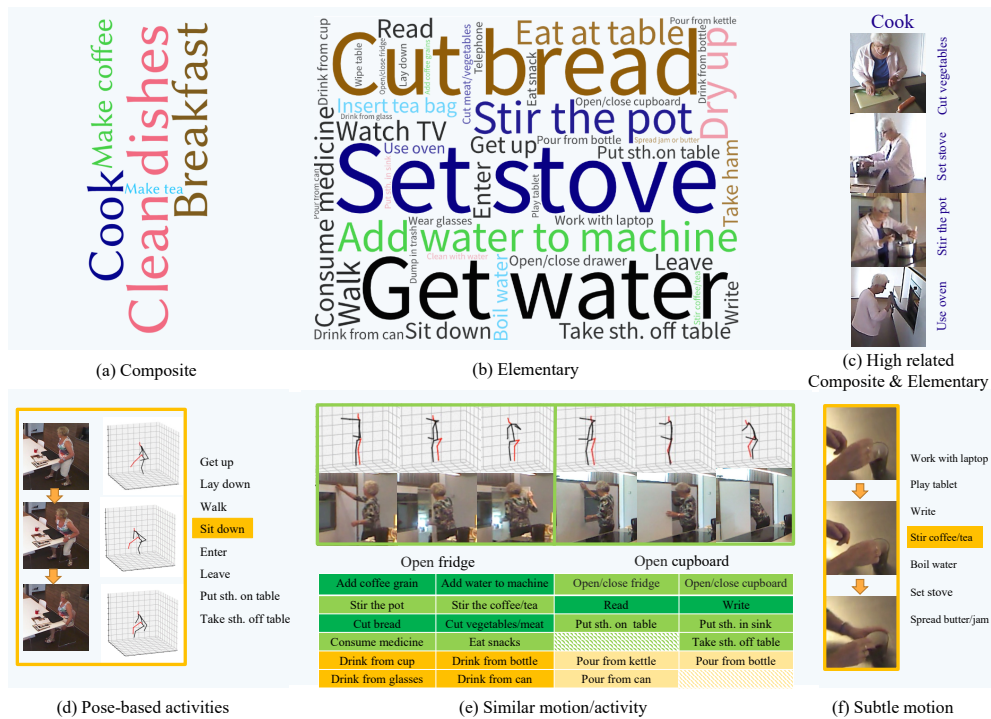

Fig. 4: On the top row, we divide the 51 activities in TSU into (a) composite and (b) elementary activities. Then, we analyze the activities along four properties: (c) highly related composite and elementary activities, (d) pose-based activities, (e) similar motion/activities, and (f) activities with subtle motion.

### 3.3.3 Challenges

TSU provides the 7 real-world challenges which are discussed in Section 2.1. (1) **Spontaneous behaviour**: TSU is an untrimmed ADL dataset where people are recorded while performing activities in a spontaneous manner. This property defines the uniqueness of TSU dataset. (2) **Low camera framing**: because of the long duration of the recording, the subjects do not pay attention to the fixed cameras. Therefore, activities can be performed very far, very close or out of view of the camera. Activities can also be partially occluded by furniture. (3) **Object-based activities**: The annotations in TSU include the fine-grained details of activities performed using different objects (e.g. *drinking from a cup, can or bottle*). TSU contains 7 object-based activities. (4) **Multi-views**: TSU features 7 camera views. As shown in Fig. 5, the camera placement enables 2-3 camera views for each environment. In this work, we use these different views for increasing the view diversities in order to design view-invariant methods and also use them for joint-view action detection (see Sec. 3.3.4). (5) **Composite activities**: TSU contains 5 composite activity classes and 16 related elementary activity classes. (6) **Concurrent activities & dense annotation**: TSU contains up to 4 concurrent activities for a single frame. About 10% of the frames contains more than one activity label. On an average, there are about 76 activity instances per video. (7) **High temporal variance**: This new dataset offers a large variation of activity duration and intra-class temporal variance. TSU features short activities (e.g. *taking on glasses*), long activities (e.g. *reading book*), and instances of the same class that can be long or short (e.g. *writing* ranges from 3 seconds to 10 minutes). As a result, handling temporal information is critical to achieve good detection performance on TSU.

### 3.3.4 Balanced TSU and Joint-view TSU

To address the different interests pertaining to long untrimmed videos, besides the aforementioned fine-grained version of the dataset (dubbed Fine-grained TSU), we introduce two additional versions of the dataset: (1) Balanced TSU and (2) Joint-view TSU. Each dataset version has its specific videos, annotations and evaluation protocol.

**Balanced TSU** is a version of the dataset that overlooks the fine-grained details (e.g., the manipulated object) but keeps only the different movement patterns (e.g., *cut*, *drink*). There are many activity classes that have a limited number of instances (i.e., samples) in the fine-grained TSU version. This is because some activity classes with specific fine-grained details occur rarely as activity instances within the dataset which may not be sufficient to learn activity-specific representations. Thus to handle this, we release Balanced TSU, which focuses on the different movement patterns of the activity (i.e., verb) rather than the fine-grained details (i.e., noun). Balanced TSU shared the same untrimmed videos as fine-grained TSU: 536 videos with 21 minutes average duration. The only difference lies in the annotation. This version of the dataset merges the fine-grained activities that share similar motion into the same activity class (e.g., *cut*

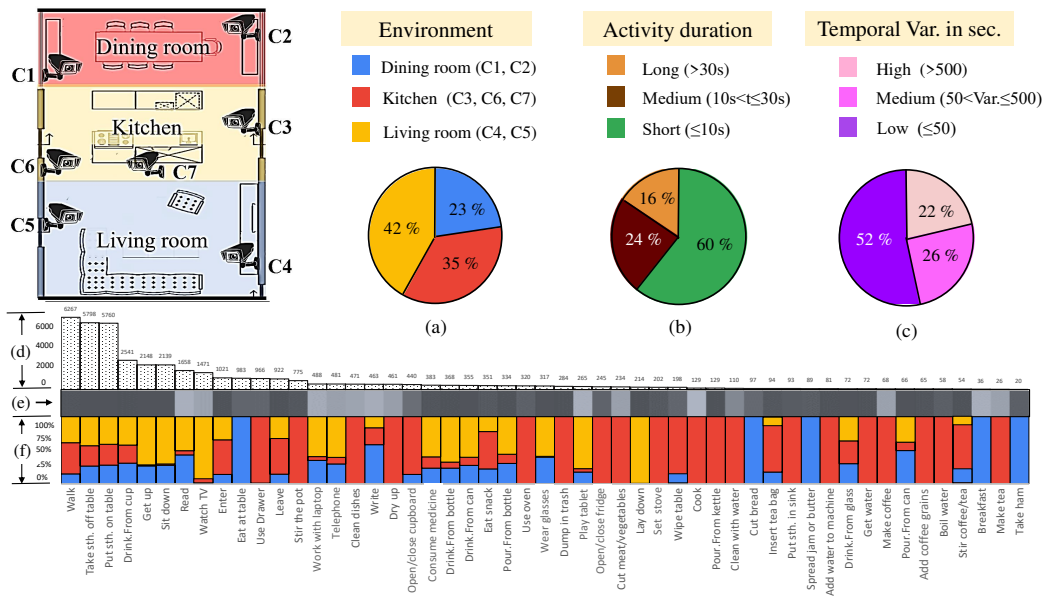

Fig. 5: On `top` row (from left to right): we provide the 7 camera locations (C: camera); activity distribution along the different (a) environments, (b) duration and (c) temporal variance. Remark: (a) is per activity instance, (b),(c) are per activity class. On `bottom` row: we provide the (d) instance frequency and corresponding (e) temporal variance heat map (e.g. the lighter the larger variance), (f) distribution of performing environment for each activity.

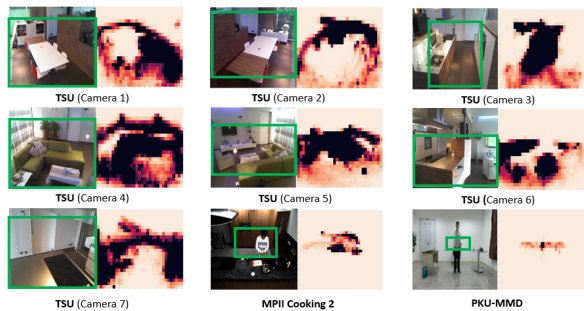

Fig. 6: Spatial Distribution of the person location in normalized image coordinates for 3 datasets, dark regions correspond to high frequency areas of the person position. The green bounding boxes embrace the high frequency locations. From the size of the bounding box, we find that TSU exhibits the largest spatial scatter, indicating the low camera framing property.

*bread*, *cut vegetables* → *cut*). Therefore, this version of dataset is more balanced in terms of the number of samples, with slightly less number of classes (in total 34 activity classes). The activity list and instance frequency are provided in Supplementary material.

**Joint-view TSU** targets the joint-view activity detection task. Different from the aforementioned versions, this version of dataset contains only synchronized video pairs to be used for learning joint-view activity detection models. We have collected 177 synchronized video pairs from the original recording footage. Each video pair contains the video content recorded from two different cameras (i.e., Dining room: C01, C02; Kitchen: C03, C06; Living room: C04, C05). For the annotation, this Synchronized version features fine-grained activity annotation. In the aforementioned fine-grained TSU, we have annotated each video separately by different annotators, resulting in potentially different annotations of the synchronized activity pairs. In particular, the synchronization can cause a shift in the activity boundaries. To build the Joint-view TSU version, the annotators have manually modified and adjusted the annotations for those synchronized videos.

All three versions will be provided in the TSU dataset website along with the annotation and videos for each version.

### 3.3.5 Toyota Smarthome Trimmed Vs Untrimmed dataset

The Toyota Smarthome Trimmed dataset contains only a single activity instance per video. In contrast, TSU dataset is composed of untrimmed videos and these videos are intermixed with multiple activity instances and backgrounds. The complexity of the problem is increased by the presence of **concurrent activities** and **composite activities**. Learning the dependencies across such activity instances is an important prospect for video understanding which was not considered in the previous trimmed version of Smarthome. Both the trimmed version and TSU feature **spontaneous behaviours**. As untrimmed videos contain multiple activities, the degree of spontaneity is also enhanced by the dependencies among the activities. For example, with spontaneous behaviour, the order of the elementary activities in composite activities can vary largely in untrimmed videos. For **intra-class temporal variance**, activity recognition methods on trimmed videos can handle this issue easily by sampling a fixed number of frames from different videos. However, in untrimmed videos where the task involves predicting the activity occurring at each timestamp, sampling mechanisms could lead to imprecise detection of activity boundaries.

TABLE 2: Comparison between the two versions of Toyota Smarthome.

| Dataset Version | Smarthome Trimmed [16] | Smarthome Untrimmed |
|---|---|---|
| Task | Recognition | Localization |
| #Classes | 31 | **51** |
| #Instances | 16 K | **41 K** |
| #Frames | 3.9 M | **13.8 M** |

Thus, learning an activity classifier for untrimmed videos which is robust to intra-class temporal invariance is a real-world challenge and is often ignored in trimmed scenarios. Concerning **data size**, as shown in Table 2, TSU is 1.6 times larger in terms of activity classes compared to the previous version of the dataset, 2.8 times larger in terms of activity instances, and 3.5 times larger in terms of total number of frames.

### 3.3.6 Benchmark Evaluation

In TSU, we define 2 evaluation protocols: *Cross-Subject* and *Cross-View*. We provide also two evaluation metrics (frame-based and event-based mAP). For frame-based evaluation, we adapt the protocol of [55] to evaluate the same mAP metric on single frames. This way of evaluating detection is robust to annotation ambiguity. For event-based evaluation, we adapt the protocol of [14]. This metric enables us to get a better insight into activity detection as not biased by activity duration.

**Cross-Subject (CS):** For cross-subject evaluation, we split the 18 subjects into training and test sets. To balance the number of videos for each activity category, we use 11 subjects for training and the 7 remaining ones for testing. This protocol considers all the 51 activities.

**Cross-View (CV):** For cross-view evaluation, the training set contains the videos from cameras 1, 3, 4, 6, 7. The remaining cameras (2, 5) are reserved for testing. The training set contains all the 51 activities and the testing set contains 32 activities from these two camera views.

## 4 THE PROPOSED BASELINE METHOD

To address the challenges in TSU dataset, we introduce an end-to-end baseline method: Attention Guided Net (AG-Net) for activity detection which is built upon temporal convolutional networks [41]. An overview of the AGNet is shown in Fig. 7. The input is the encoding of a video. The AGNet has two principal components: a stacked dilated temporal convolution network (SD-TCN) and an attention module. In this work, the input to the base-network is always the RGB frames. For attention module, the input is another modality, such as 3D human poses or optical flow. For simplicity, in the following, we consider the 3D poses as the input to the attention module. The SD-TCN and the attention module have both a 5-block structure. These blocks have temporal convolution with increased dilation rates setting, thus the receptive field increases exponentially. The lower-blocks have smaller reception fields while the higher blocks have larger receptive fields. For every block, the pose-attention module generates an attention mask that represents the temporal saliency of human activities in a video. The main contribution is the attention module, which utilizes 3D poses to generate the attention weights at multiple temporal scales. We believe that 3D poses are complementary to the RGB modality as they help filtering the irrelevant context in the RGB frames and providing more weight to the pertinent frames of the video.

Below, we detail the video encoding and the model structure of AGNet.

### 4.1 Video encoding

Similar to most activity detection models [41], [42], [47], our model processes the encoding of video segments. In this work, we use state-of-the-art convolution model (i.e. 2D+T CNN or (2+1) D+T GCN) to extract appearance features in the video. The RGB encoding is extracted by a CNN such as Inception [44] or I3D [33]. The pose encoding is extracted by a GCN such as ST-GCN or 2s-AGCN [37]. We fine-tune the 3D convolution model on the training set of TSU to better model the spatial information in this dataset.

**Training:** To fine-tune the feature extraction model, firstly, we divide the video into 100-frame-long non-overlapping segments. For the RGB modality, to tackle the camera framing challenge, we apply SSD [56] to extract the human crops (i.e. bounding box) of the subject, and resize the crop into 224×224. For 3D poses, the subject would always be re-projected at the center of the screen with a fixed scale by using [49]. We then train the classification model [33], [37] with the uni-sampled 16 frames for each segment. For the RGB modality, we flip all the images in each segment with a probability of 0.5. The inputs to the RGB or 3D pose convolution model are the RGB human crops and corresponding skeleton of a segment respectively. We optimize the multi-label binary cross-entropy loss [57] to learn the parameters.

**Feature extraction:** To extract the features, a video is divided into $T$ non-overlapping segments, each segment consisting of 16 frames. These segments of RGB human crops or pose sequences are sent to the fine-tuned spatio-temporal model to extract the segment representation. We stack the segment-level features along the temporal axis to form a $T \times C_{in}$ dimensional video representation where $1 \times C_{in}$ is the feature shape per segment. This video representation denoted as $F_{in}$ is further input to the RGB or pose stream in our architecture.

### 4.2 Model structure

In this section, we present the structure of the AGNet.

Our stacked-dilated temporal convolution network (SD-TCN) is a TCN-based network. This network has 5 blocks, each block has one 1-dimensional convolution layer, one Hadamard product with the attention weights from the attention module and a residual link. For different blocks, we give different dilation rates to the convolution layer. With these different settings in dilation, we can model local context in the lower block and global context in higher blocks. In our experiment, we set the kernel size ($k$) to 3 for all convolution layers, dilation ($d_i$) and padding rate to $2^{i-1}$, thus the reception field is up to $2^i + 1$ for the $i^{th}$ block.

In parallel to the SD-TCN, the attention module is another TCN-based model. The attention module has a similar 5-block structure as the SD-TCN, and also the same kernel and dilation setting for the convolution inside the block. Thus, the attention module has the same receptive field as the SD-TCN for each block. However, this module uses significantly lower channel capacity to generate the attention weights. For each convolution layer, it has a ratio of $\beta$ ($\beta \leq 1$) channels for the SD-TCN. The typical value is $\beta$ = 1/8 in our experiments, which is much lower than the SD-TCN. In the attention module, after the convolution layer, we generate the attention map $A_i$. A bottleneck layer is applied as a transformation to match the channel size to the SD-TCN. Normalizing the high number of T attention weights with softmax leads to extremely low values, which can hamper their effect. To avoid this, we use sigmoid activation to generate the final attention map.

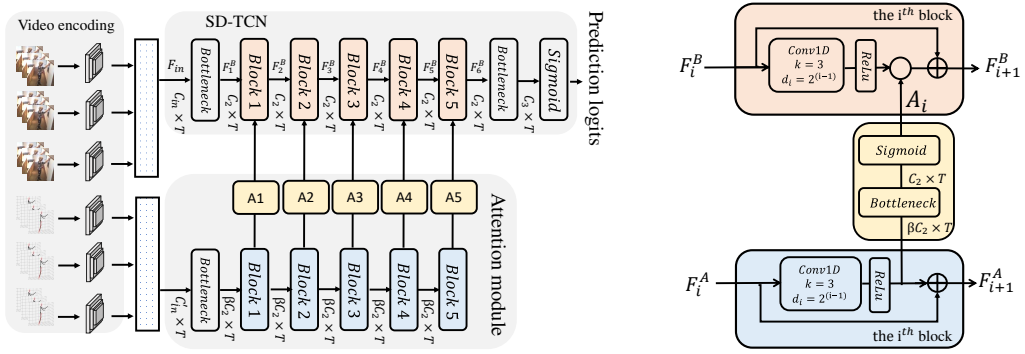

Fig. 7: On the left, we present the overview of the AGNet. In this figure, `Bottleneck` indicates the 1D convolution that processes the features across time and which kernel size is 1. On the right, we present the computation flow for one block. In each block, `k` is the kernel size and `d` is the dilation.

As shown in Fig. 7, the input RGB and pose encoding are firstly fed to the bottleneck layers. The output channel size from the bottleneck layers is $C_2$ and $\beta C_2$, corresponding to the SD-TCN and attention module respectively. Then 5 blocks are stacked, the set of operations in each block can be formulated as follow:

$$F_{i+1}^A = F_i^A + ReLU(Conv1D(F_i^A, k, d_i)) \qquad (1)$$

$$A_i = Sigmoid(W_i ReLU(Conv1D(F_i^A, k, d_i))) \qquad (2)$$

$$F_{i+1}^B = F_i^B + ReLU(Conv1D(F_i^B, k, d_i)) \circ A_i \qquad (3)$$

where $F_i^B$ and $F_i^A$ indicates the input feature map of the $i^{th}$ block of the SD-TCN and attention module respectively. $A_i$ is the attention mask generated from the $i^{th}$ block. $\circ$ indicates the Hadamard product. $W_i \in \mathbb{R}^{C_2 \times \beta C_2}$ are the weights of the bottleneck convolution in attention module.

Finally, we compute the per-frame binary classification score for each class (i.e. prediction logits). The classifier is on top of the SD-TCN, which is another bottleneck convolution with *sigmoid* activation:

$$P = Sigmoid(W' F_6^B) \qquad (4)$$

where $P \in \mathbb{R}^{T \times C_3}$ are the prediction logits and $W' \in \mathbb{R}^{C_3 \times C_2}$ are the weights of the bottleneck convolution, $C_3$ corresponds to the number of activity classes. To learn the parameters, we optimize the multi-label binary cross-entropy loss [57].

## 5 EXPERIMENTS

The goal of these experiments is to verify that the TSU dataset provides novel challenges that are not yet addressed by the other state-of-the-art datasets. For that, we show that the state-of-the-art detection methods perform poorly on TSU and that our AGNet significantly improves the results on TSU as it is designed to address the targeted real-world challenges. To evaluate the effectiveness of the AGNet, we compare it on TSU dataset with 9 detection methods, which represent the state-of-the-art on other densely-annotated datasets [9], [13]. We also perform a comparative study between TSU and the challenging Charades dataset for the activity detection task to better highlight how real-world challenges are addressed by both datasets.

### 5.1 Implementation details

#### 5.1.1 Video encoding

We use three types of encoders to extract the encoding of the input videos. As described in section 4.1, AGCN [37] and I3D [33] are fine-tuned on TSU and then the features are extracted. Moreover, we also evaluate this dataset on per-frame features. We use Inception V1 [44] pre-trained on ImageNet [58] to extract the features. The channel size of I3D and Inception is 1024, the channel size of AGCN is 256.

#### 5.1.2 State-of-the-art methods

Nine activity detection methods are evaluated on our dataset: bottleneck, Non-local network [39], LSTM [59], Bidirectional-LSTM [60], Dilated-TCN [41], R-I3D [34], Super event [42], TGM [47] and MS-TCN [43]. Bottleneck has only one dropout layer (with dropout probability 0.5) and a bottleneck layer as the classifier. Non-local [39] has one non-local block applied on the features of the whole video before the classifier. LSTM [61] has one LSTM layer with 512 hidden units and one dropout layer (with dropout probability 0.5). Similarly, for Bidirectional-LSTM [60], we have two opposite direction 512 hidden units LSTM layers. The features are concatenated before the classifier. R-I3D [62] uses I3D [33] as its base network. We set the anchor scale value to [0.3, 0.6, 1.0, 1.5, 2, 2.5, 2.75, 3, 3.5, 4, 4.5, 5, 5.5, 6, 6.5, 7, 7.5, 8, 10, 12, 14, 16, 18, 20, 24, 28, 32, 38, 42, 50, 58, 66, 78, 84, 90, 96]. For TGM [47], we add one layer

to have a 4-layer structure. All the methods use the same video encoding as the AGNet and they are trained with binary cross-entropy loss with sigmoid activation [57]. The unspecified parameters are similar to the original papers.

*5.1.3 AGNet*

We set $N = 6$ blocks. For I3D and Inception features, the channel size is 1024, for AGCN pose features, the channel size is 256. $C_1$ is 512 and $\beta$ is 8. We use Adam optimizer [63] with an initial learning rate of 0.001, and we scale it by a factor of 0.3 with a patience of 10 epochs. The network is trained on a 4-GPU machine for 300 epochs with a mini batch of 32 videos for Charades and 2 videos for TSU.

## 5.2 Comparative study on TSU

Table 3 provides the results of the considered activity detection methods on the fine-grained version of TSU. To be noticed, the Bottleneck used for comparison is a Bottleneck on top of the segment-level features. Unlike the other methods, Bottleneck does not have further temporal processing after the video encoding part. Thus, this method cannot model long temporal information, which is crucial for activity detection. In contrast, the other activity detection baselines focus on the temporal processing. The improvement over the Bottleneck reflects the effectiveness of modeling temporal information.

TABLE 3: Per-frame mAP (%) on the Fine-grained TSU dataset.

| | | | CS | CV |
|---|---|---|---|---|
| Pose | 3D+T | AGCN+Bottleneck [37] | 10.1 | 12.6 |
| | | AGCN+LSTM [61] | 17.0 | 14.8 |
| | | AGCN+**SD-TCN** | **26.2** | **22.4** |
| RGB | 2D | Inception+Bottleneck [44] | 11.5 | 5.2 |
| | | Inception+LSTM [61] | 13.2 | 5.3 |
| | | Inception+**SD-TCN** | **22.3** | **12.1** |
| | 2D+T | R-I3D [62] | 8.7 | - |
| | | I3D(Trimmed)+Bottleneck [33] | 7.4 | 4.3 |
| | | I3D+Bottleneck [33] | 15.7 | 9.2 |
| | | I3D+Non-local block [39] | 16.8 | 9.6 |
| | | I3D+Super event [42] | 17.2 | 10.9 |
| | | I3D+LSTM [64] | 22.6 | 12.9 |
| | | I3D+Bidirectional-LSTM [60] | 24.5 | 15.1 |
| | | I3D+Dilated-TCN [41] | 25.1 | 13.9 |
| | | I3D+MS-TCN [43] | 25.9 | 13.1 |
| | | I3D+TGM [47] | 26.7 | 13.4 |
| | | I3D+**SD-TCN** | **29.2** | **18.3** |
| RGB+Pose | | **AGNet** | **33.2** | **23.2** |

*5.2.1 Ablation & Data modality on Fine-grained TSU*

In this section, we conduct the ablation and data modality analysis on the fine-grained version of the TSU. In Table 3, we firstly compare the three different video encodings: AGCN pose features, inception RGB features and I3D RGB features. We conduct the experiments on the Bottleneck, LSTM and the AGNet. The AGNet is the SD-TCN (RGB) guided by a attention module (pose). On one hand, we observe that using I3D RGB features improves the detection results by up to 11.1% w.r.t. the same method using Inception features. This improvement is intuitive because of the higher ability of the 3D convolutional operations to capture spatio-temporal relations using several datasets for pre-training. On the other hand, we find that, while using the same method, 2D+T RGB features perform better than pose features in Cross-Subject protocol. However, pose features perform better than RGB features in Cross-View protocol (+4.1% for SD-TCN). This reflects that 3D skeleton is more stable while changing viewpoints, which is very helpful in multi-view settings as in TSU. Finally, for the AGNet: SD-TCN (RGB) guided by pose attention module, outperforms RGB and pose SD-TCNs for both CS and CV protocol (+4.0% and +4.9% w.r.t RGB SD-TCN for CS and CV protocol respectively).

TABLE 4: Event-based mAP (%) for different IoU thresholds for the Fine-grained TSU dataset. The AGNet utilizes both pose and RGB modalities and the other methods utilize only RGB.

| | CS | | | CV | | |
|---|---|---|---|---|---|---|
| IoU Threshold ($\theta$) | 0.3 | 0.5 | 0.7 | 0.3 | 0.5 | 0.7 |
| Bottleneck [33] | 5.0 | 2.5 | 0.5 | 2.3 | 1.1 | 0.2 |
| Non-local block [39] | 4.9 | 2.2 | 0.6 | 1.6 | 0.7 | 0.1 |
| Super event [42] | 5.7 | 2.8 | 0.7 | 1.8 | 0.9 | 0.1 |
| LSTM [61] | 11.6 | 6.4 | 2.2 | 6.0 | 3.2 | 0.7 |
| Bidirectional-LSTM [60] | 13.3 | 7.9 | 3.5 | 9.0 | 5.4 | 1.2 |
| Dilated-TCN [41] | 12.8 | 6.9 | 3.0 | 5.8 | 3.3 | 0.8 |
| MS-TCN [43] | 13.2 | 7.6 | 3.0 | 5.3 | 3.1 | 0.4 |
| TGM [47] | 15.1 | 9.4 | 4.2 | 5.5 | 3.2 | 0.4 |
| **AGNet** | **22.7** | **15.3** | **6.0** | **12.5** | **7.8** | **2.9** |

We then show that the method trained on the trimmed version (i.e. I3D(Trimmed)+Bottleneck) fails to generalize to the untrimmed version. Firstly, we train an I3D model with the trimmed version of TSU (51 class version). Secondly, we leverage a sliding window framework to utilize the I3D model to predict the activity class for each window, in which the classifier is fine-tuned for the frame-level activity detection task. Note that I3D (Trimmed)+Bottleneck is very close to the I3D+Bottleneck model. The difference mainly lies in the I3D training process. For this baseline I3D (Trimmed)+Bottleneck, I3D is trained with the clipped activity instances, whereas I3D + Bottleneck is trained with random snippets that may include activity instances or even a mix of activities and background. From Tab. 3, we find that this baseline trained on the trimmed version underperforms in detecting activities in TSU. This is due to the lack of contextual relationships present among the activity instances in the trimmed version and hence the baseline fails to generalize over the untrimmed scenario.

Inspired by Charades [13], to understand the relation between the number of activity samples and performance, Fig. 8 illustrates AP for each activity. In this figure, the activity classes are sorted by the number of available samples, together with the name of best performing classes. The number of samples in a class is primarily decided by the universality of the activity (can it happen in any scene), and if it is typical of household environments. It is interesting to notice that, while there is a trend for activities with a higher number of examples to have higher AP, it is not true in general. Activities such as *breakfast*, and *get water* have top-10 performance despite being represented by only few examples.

To understand the advantages of 3D skeleton and RGB modality, in Fig. 9, firstly, we select the top 10 activities where 3D skeleton stream outperforms RGB stream in CV protocol. We find that 5 out of the 8 pose-based activities that we defined in Fig. 4 (4) are in these top 10 activities. This confirms that 3D skeleton stream has filtered the unnecessary context information in the image, resulting in a better model for the posed-based activities. Secondly, we select the top 10 activities where RGB stream outperforms 3D skeleton stream in CS protocol. We find 7 out of 10 activities are the similar activities with different objects that we defined in Fig. 4 (5). This confirms that RGB stream provides the object information lacking in 3D skeleton, which is critical to detect the activities highly correlated with objects. Finally, we show that, while using our attention-based baseline, we can handle both challenges of pose-based activities and similar activities involving different objects.

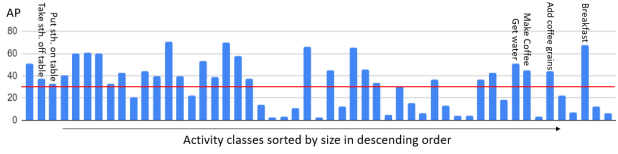

Fig. 8: Average Precision for the activities in Fine-grained TSU. The classes are sorted by their size. The mAP is marked by a red line. We can see that while there is a slight trend for smaller classes to have lower accuracy, many classes do not follow that trend.

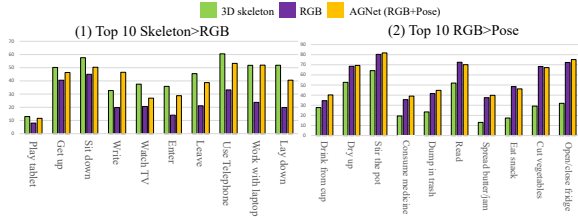

Fig. 9: Frame-based mAP of the AGNet using different modalities: (1) Top 10 activities where the 3D skeleton stream outperforms the RGB stream for the CV protocol. (2) Top 10 activities where the RGB stream outperforms the 3D Skeleton stream for the CS protocol.

In Fig. 10, we present the attention map of the attention module for 5 layers (on top), and the corresponding ground truth vs. activity detection results (on the bottom). On the one hand, in area (A), while detecting short activities, the attention module allocates high attention weights at the lower layer, corroborating that the lower layer is particularly sensitive to short activities. On the other hand, in area (B), with long activities (e.g. *Read book*), only the higher layers allocate high attention weights to the frames in the kernel. This reflects that the higher layers are more sensitive to long-term activities.

In summary, we find that the available modalities in TSU are complementary. The AGNet leverages these modalities to address the challenges in TSU such as multi-views, pose-based activities and similar motions.

#### 5.2.2 Analysis on Balanced TSU

For Balanced TSU, we evaluate the performance of SD-TCN in per-frame mAP with Cross Subject split. In table 5, we find that for the three baseline methods, the model trained with the Balanced annotation can achieve higher performance compared to the fine-grained version. This en-courages the network to learn the pertinent motion patterns relevant for an action rather than the fine-grained details (like low resolution objects involved) from small number of action instances.

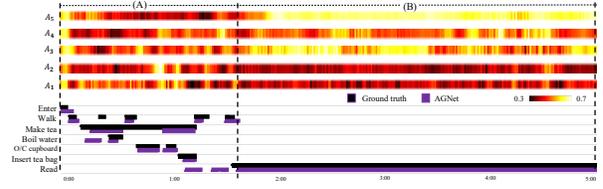

Fig. 10: Qualitative analysis of the detection result and the attention map. On the top, we visualize the attention map $A_i$ for 5 layers. On the bottom, we present the corresponding ground truth and detection performance for an example video.

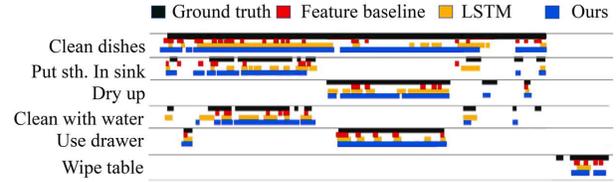

Fig. 11: Qualitative study

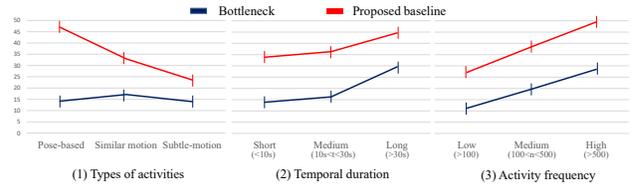

Fig. 12: We compare the AGNet against the Bottleneck approach across three different activity properties using both RGB and Pose modality. Evaluation is provided on frame-based mAP on TSU-CS. The Bottleneck performs poorly on all these types of activities, whereas the AGNet improves the performance on all of them.

#### 5.2.3 Analysis on Joint-view TSU

With the synchronized video pairs, we study the effect of accessing synchronized multi-view videos towards improving the activity detection performance. Firstly, we utilize I3D to extract the spatio-temporal features of the videos, then SD-TCN is applied on top of those I3D features to perform activity detection. View 1 and View 2 indicate the two views for the synchronized videos. Figure 13 shows the experimental design. On the top, we show the baseline of the **mixed-view activity detection**: SD-TCN is trained with a mixture of the synchronized videos for the activity detection task for both training and testing. On the bottom, we show the **joint-view activity detection** baseline: both videos from the synchronized video pairs are fed to two SD-TCNs, then the prediction scores are fused to have the prediction for combined views. The number of training videos is the same as the mixed-view activity detection. The difference is that the joint-view baseline inputs two synchronized videos at a time. Moreover, we adopt the proposed baseline AG-Net for the joint-view activity detection. In this experiment, the inputs to AG-Net are the synchronized videos with the same modality. In table 6, we find that training the network with joint-views can boost the performance with a large margin (+ 3.4%, +4.6% w.r.t. mixed-view baseline), especially for the AG-Net baseline.

TABLE 5: Balanced TSU. Evaluation for the cross-subject split with per-frame mAP.

|  | Fine-grained TSU | Balanced TSU |
|---|---|---|
| I3D+Bottleneck [33] | 15.7 | 17.1 |
| I3D+TGM [65] | 26.7 | 29.0 |
| I3D+SD-TCN | 29.2 | 35.8 |

TABLE 6: View 1 and View 2 indicate the two views for the synchronized videos. Different settings are evaluated by SD-TCN with per-frame mAP in Cross-Subject split.

| Setting | Input | mAP |
|---|---|---|
| Mixed view | View 1, View 2 | 35.5 |
| Joint View - Late Fusion | View 1, View 2 | **38.9** |
| Joint View - AG-Net | View 1, View 2 | **40.1** |

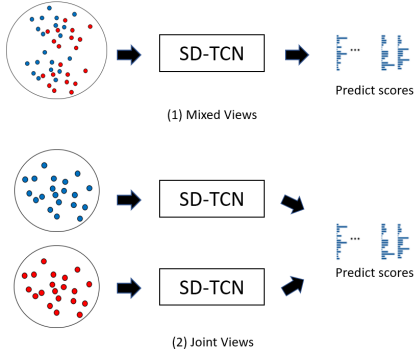

Fig. 13: Joint-view activity detection. Each colorful small dots indicates an I3D feature representation of an untrimmed video. A red and a blue dots form a pair of synchronized videos.

### 5.2.4 State-of-the-art comparison

In Table 3, we compare the performance of the 9 considered methods on Fine-grained TSU. The comparative study is conducted with the I3D RGB features. The first method is a proposal-based method that adopts R-C3D [62] with I3D base network (we call this method R-I3D). This method fails to generate precise proposals for long activities with dense labels due to high computational cost. Consequently, it yields the worst detection performance on Fine-grained TSU. The second and the third methods are the Bottleneck [33] and the Non-local block [39]. We find that the non-local block can provide the information of one-to-one temporal dependency to the local features (+ 0.9% w.r.t. Bottleneck on TSU-CS), however, Non-local block is not effective enough. Similarly, Super-event [42] utilizes temporal structure filters to model latent representation of composite activities and then compute their affinity with each frame (+ 4.2% w.r.t. Bottleneck on TSU-CS). However, videos in Fine-grained TSU are long and complex, thus it is hard to model latent representation of composite activities in this dataset. We need the temporal filter to gradually embed the information of the local frames to the current frame. LSTM [61] and Bidirectional-LSTM [60] are RNN based methods. These methods can model short temporal relations (up to +8.8% w.r.t. Bottleneck on TSU-CS), but fail to model the long temporal relationships in the complex activities of TSU. Dilated-TCN [41], TGM [47], MS-TCN [43] use temporal Gaussian/Convolutional filters which better capture the temporal relationships in long activities (up to +13.5% w.r.t. Bottleneck on TSU-CS). Thanks to the effect of temporal filters, these methods can process long-term temporal relations. Similarly, our proposed method leverages 3D poses to generate the attention weights at multiple temporal scales, which help to detect the activities with variable temporal length and from multi-views. As a result, the AGNet outperforms all the state-of-the-art methods by a significant margin (+17.5% w.r.t. Bottleneck on TSU-CS).

In Fig. 11, we show qualitative visualization results of three model predictions. In this video, there are one composite long activity and 5 elementary activities. We notice that our AGNet can better tackle the long-term temporal relations, detecting the composite activity and the related elementary activities simultaneously. Additionally, the AGNet provides better detection for both elementary (e.g. *wipe table*) and composite activities (e.g. *Clean dishes*) compared to I3D and LSTM. However, the detection precision is not sufficient, more work is needed to design better models to detect both composite and elementary activities in untrimmed videos.

In Fig. 12, we compare the performance across 4 different activity properties of the AGNet and Bottleneck using both RGB and pose modalities (i.e. I3D+AGCN). Bottleneck layer is the baseline reflecting the quality of the feature without temporal processing. Thus, the comparison with the Bottleneck can reflect the improvement from our proposed methods and the remaining open issues on Fine-grained TSU. In Fig. 12 (1), we observe that the AGNet significantly improves the detection of pose-based activities compared to Bottleneck. However, the AGNet does not tackle so well similar motion and subtle motion activities. In Fig. 12 (2), we show that longer activities are easier to recognize than shorter ones, similarly to [66]. The consistent performance gain of the AGNet for activities with different temporal duration corroborates its effectiveness to adapt to temporal dynamics. Finally, we show for the AGNet the improvement in the detection of all activities, even of the ones with small numbers of training samples. We are not applying specific measures in the AGNet to handle this issue. Adopting strategies like class-weighting, optimizing through focal loss could be explored in future work.

In table 4, we present the event-based evaluation of the detection methods. The AGNet provides more precise predictions than the state-of-the-art methods. However, all these performances are relatively low, indicating that current methods are far from addressing real-world conditions.

### 5.3 Comparative analysis between TSU & Charades

The results of the activity detection methods on different datasets provide us valuable insights into the key properties of the datasets themselves. Closely related to TSU, we choose the Charades dataset to perform a comparative study. Both datasets focus on daily living activities. They are densely annotated containing many concurrent activities and object-based activities. However, these datasets differ on several points. (1) In Charades, due to the self-recorded video settings, the activities are fast and the camera framing is high, and as a consequence, the subject is always in the center of the camera view. In contrast, in TSU, the subjects performing the activities have high spontaneity leading to higher intra-class variability and lower camera framing (see Section 2.1). (2) In Charades, the larger number of activity classes originates from the combination of only 33 verbs with different objects (e.g. "holding some food", "holding a sandwich"). In comparison, the 51 activities in Fine-grained TSU originate from 35 different semantic verbs. Therefore, the Charades dataset has more activity classes relative to objects while having less semantic verbs of daily living activities. (3) TSU has longer videos (20 mins on average), compared to the on average 30 second clips in Charades. As a result, Charades does not have long activities, and the temporal variance of activity instances is low in this dataset. Fig. 15 presents the temporal duration of activity instances in Charades and Smarthome. We find that Smarthome has larger scope and higher temporal variance for the activity duration.

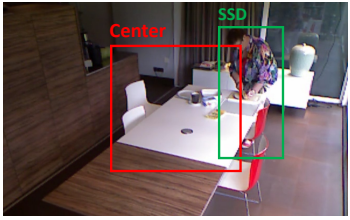

Fig. 14: SSD & Center crops

TABLE 7: Address the camera framing challenge

|  | Fine-grained TSU-CS | | Charades | |
|---|---|---|---|---|
|  | Human | Center | Human | Center |
| I3D + Bottleneck [33] | **15.7** | 10.8 | 15.8 | 15.6 |
| I3D + Super event [42] | **17.2** | 12.1 | 18.4 | 18.6 |
| I3D + AGNet | **33.2** | 26.9 | **22.9** | 22.8 |

To first quantify the level of camera framing in TSU as compared to Charades, we evaluate three baseline methods trained/tested using crops around the human body or crops in the middle of the images (Fig. 14). The crops around the human body are extracted using SSD [56]. The results are reported in Table 7. To evaluate the performance on Charades, we measure the frame-based mAP for activity detection [42], [47], [55] using the settings described on the dataset's website. For Charades, the methods using human crops and center crops obtain similar results, suggesting that Charades has high camera framing—that is, the subject in the videos is usually centered within the frames. On the other hand, in TSU, the use of human crops improves performance significantly (nearly +6%). Indeed, TSU has low camera framing—that is, subjects often perform activities at the image borders.

We also compare our Attention Guided Net with the state-of-the-art on Charades. Note that, as poses are not provided, we utilize optical flow as the input of the at-tention module. Table 8 shows that our proposed method outperforms the other methods (+ 0.6% w.r.t. TGM + Super event using both RGB + Flow). However, due to the different properties between TSU and Charades, the improvement of our multi-temporal scale attention mechanism on Charades is not as significant as on TSU. Firstly, the unnatural low variance of the activity duration in Charades limits the need for the multi-temporal scale structure of SD-TCN. Secondly, there are many activity classes with a similar motion with different object information in Charades. The attention module utilizes complementary modality information to guide the SD-TCN. The pose or optical flow can provide salient motion patterns, but not the object details, which limits the improvement yielded by the attention module.

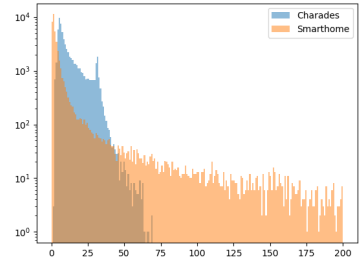

Fig. 15: Histogram of activity instance duration in Smarthome and Charades. X axis represents the duration, Y axis represent the number of instances in log scale.

To summarize, both Charades and TSU are challenging datasets. Their challenges are complementary. Charades dataset focuses more on the environment diversity, whereas TSU focuses on the spontaneity of activities (increasing the intra-class variance) and on the temporal relationships between composite and elementary activities (increasing the activity complexity).

TABLE 8: Per-frame mAP (%) on Charades, evaluated with the Charades localization setting. Note: cited papers may not be the original paper but the one providing this mAP results.

|  | Modality | mAP |
|---|---|---|
| Two-stream [55] | RGB + Flow | 8.9 |
| Two-stream+LSTM [55] | RGB + Flow | 9.6 |
| R-C3D [62] | RGB | 12.7 |
| Asynchronous Temporal Fields [55] | RGB + Flow | 12.8 |
| I3D [42] | RGB | 15.6 |
| I3D [42] | RGB + Flow | 17.2 |
| I3D + 3 temporal conv.layers [47] | RGB + Flow | 17.5 |
| TAN [67] | RGB + Flow | 17.6 |
| I3D + WSGN (supervised) [68] | RGB | 18.7 |
| I3D + Stacked-STGCN [69] | RGB | 19.1 |
| I3D + Super event [42] | RGB + Flow | 19.4 |
| I3D + 3 TGMs [47] | RGB + Flow | 21.5 |
| I3D + 3 TGMs + Super event [47] | RGB + Flow | 22.3 |
| I3D + **SD-TCN** | RGB | **21.6** |
| I3D + **AGNet** | RGB + Flow | **22.9** |

## 6 CONCLUSION

In this paper, we propose the Toyota Smarthome Untrimmed dataset, a novel untrimmed video dataset, that features spontaneous behaviours and several other real-world challenges for activity detection. This dataset includes three versions: (1) Fine-grained TSU, (2) Balanced TSU and (3) Joint-view TSU. Each version addresses a specific interest regarding the targeted activity detection task. We conduct a comparative study with other activity detection datasets, and highlight the added value of TSU dataset, w.r.t. activity diversity, camera framing, dense annotation at two levels (i.e. composite and elementary activities) and high variation of activity duration. For instance, regarding activity diversity, TSU contains a balanced number of activities characterized by specific poses, by complex interactions with objects, by subtle and similar motions. Moreover, the appearance of these activities can vary a lot according to the subject behaviour, the camera view point, occlusion level and type of environment. We provide a set of statistics to better quantify TSU contributions.

We also propose a baseline method to address some of these real-world challenges. For instance, for dealing with low camera framing, we use human crops to extract features that are more relevant to the activities. For dealing with large temporal variance, the attention module generates attention masks at different temporal scales to help detect activities with different temporal lengths. For multi-view challenge, we use both RGB and 3D skeleton to better tackle the view variance problem. We show that our baseline significantly outperforms the state-of-the-art on all the evaluation protocols of TSU. We evaluate the proposed method on the Charades dataset, showing that also on this dataset, our baseline achieves state-of-the-art results, thus confirming its effectiveness. The fact that the overall performance on TSU is still rather low, indicates that some of the issues related to real-world conditions are still to be tackled and that there is still room for improvement. As future work, we plan to leverage the multiple modalities in the training phase while using only RGB at inference time. We also plan to investigate the long-tail distribution challenge. Besides, we plan to explore the relations between composite and elementary activities. For example, we envisage to train with composite activities in order to predict the elementary activities in a weakly supervised manner. The TSU dataset has just been released to public for academic research purposes[1]. This will allow researchers to develop novel approaches to promote activity detection in the wild.

1. project page: https://project.inria.fr/toyotasmarthome/

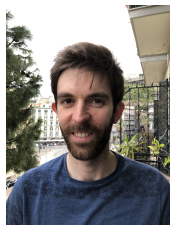

**Lorenzo Garattoni** obtained his Bachelor and Master degrees from the University of Bologna, Italy. In 2012, he started a PhD at the Université libre de Bruxelles, focusing on cognition in multi-robot systems, under the supervision of Prof. M. Birattari. Since 2018, he has been working for Toyota Motor Europe, where he is a senior engineer developing perception technologies and control software for robots.

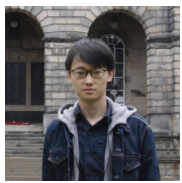

**Rui Dai** obtained his Bachelor's degree from Beihang University (BUAA) in Beijing, China. In 2018, he received his Master's degree from INP-ENSEEIHT, Université de Toulouse. Since 2018, he has been a PhD student at Inria and Université Côte d'Azur in France, under the supervision of Dr Francois Bremond. His research interests include video understanding, human-object interaction, cross-modal learning, etc. His ideas have been accepted by top-tier journals and conferences on computer vision.

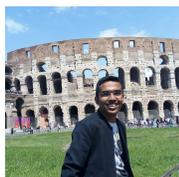

**Srijan Das** is a Postdoctoral Associate at Stony Brook University, New York. He received his Ph.D. in computer science from University Côte d'Azur and Inria Sophia Antipolis, France. His research interest includes video understanding (particularly human action recognition), knowledge distillation, and attention mechanisms. During his research career, he has proposed several pose-driven spatio-temporal attention mechanisms for human action recognition. Many of these ideas have been accepted to top-tier conferences and journals on vision such as ECCV, ICCV, WACV, PRL. He is also serving the community by reviewing and organizing scientific events.

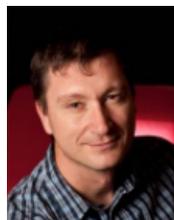

**Francois Bremond** received the PhD degree from INRIA in video understanding in 1997, and he pursued his research work as a post doctorate at the University of Southern California (USC) on the interpretation of videos taken from Unmanned Airborne Vehicle (UAV). In 2007, he received the HDR degree (Habilitation a Diriger des Recherches) from Nice University on Scene Understanding. He created the STARS team on the 1st of January 2012. He is the research director at INRIA Sophia Antipolis, France. He has conducted research work in video understanding since 1993 at Sophia- Antipolis. He is author or co-author of more than 140 scientific papers published in international journals or conferences in video understanding. He is a handling editor for MVA and a reviewer for several international journals (CVIU, IJPRAI, IJHCS, PAMI, AIJ, Eurasip, JASP) and conferences (CVPR, ICCV, AVSS, VS, ICVS). He has (co-)supervised 26 PhD theses. He is an EC INFSO and French ANR Expert for reviewing projects.

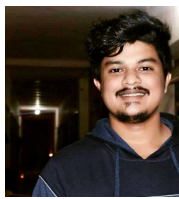

**Saurav Sharma** received his B.Tech degree in Computer Science in 2012 from Tezpur University, Assam, India and M.Tech degree in Computer Science in 2017 from National Institute of Technology (NIT) Rourkela, Odisha, India. His research interests includes but not limited to video understanding tasks. He is currently working in a full stack growth startup as a Decision Scientist and leads cloud native industrial projects in the domain of Computer Vision and NLP.

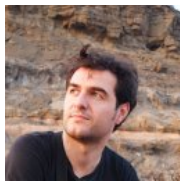

**Gianpiero Francesca** obtained his Bachelor and Master degrees from the "Sannio" University in Benevento, Italy. In 2011 he started a PhD at the ULB University in Brussels, focusing on swarm robotics and automatic design of control software, under the supervision of Prof. M. Birattari. Since 2015, he has been working for Toyota Motor Europe. Today he is a Senior Engineer there, developing human activity recognition technology.

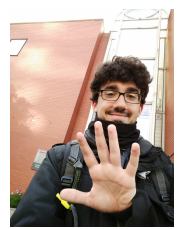

**Luca Minciullo** obtained his Bachelor and Master degrees from the "La Sapienza" University of Rome, in 2012 and 2014 respectively. In 2014, he started a PhD at the University of Manchester in computer vision and medical imaging under the supervision of Prof. T.F. Cootes. Since 2018, he is an Engineer at Toyota Motor Europe, developing perception technologies for indoor robotics.